\documentclass[sigconf]{acmart}

\usepackage{multirow}
\usepackage{pifont}
\usepackage{balance}

\AtBeginDocument{%
  }

\copyrightyear{2026}
\acmYear{2026}
\setcopyright{cc}
\setcctype{by}
\acmConference[MM '26]{Proceedings of the 34th ACM International Conference on Multimedia}{November 10--14, 2026}{Rio de Janeiro, Brazil}
\acmBooktitle{Proceedings of the 34th ACM International Conference on Multimedia (MM '26), November 10--14, 2026, Rio de Janeiro, Brazil}
\acmDOI{10.1145/3767308.3836263}
\acmISBN{979-8-4007-2213-4/2026/11}
\begin{document}

\title{SportsGrounder: Proposal-Aided Interleaved Grounding for Dense Sports Video Reasoning}


\author{Yizhi Li}
\orcid{0009-0000-1692-5778}
\affiliation{%
  \institution{Zhejiang University}
  \city{Hangzhou}
  \country{China}
}
\email{yizhi.24@intl.zju.edu.cn}

\author{Jiawei Jiang}
\orcid{0009-0009-1585-5726}
\affiliation{%
  \institution{Beijing Jiaotong University}
  \city{Beijing}
  \country{China}
}
\email{byjiaweijiang@gmail.com}

\author{Guanhong Wang}
\orcid{0009-0007-6058-5468}
\affiliation{%
  \institution{Zhejiang University}
  \city{Hangzhou}
  \country{China}
}
\email{guanhongwang@zju.edu.cn}

\author{Yingcai Wu}
\orcid{0000-0002-1119-3237}
\affiliation{%
  \institution{Zhejiang University}
  \city{Hangzhou}
  \country{China}
}
\email{ycwu@zju.edu.cn}

\author{Gaoang Wang}
\authornote{Corresponding author.}
\orcid{0000-0002-8403-1538}
\affiliation{%
  \institution{Zhejiang University}
  \city{Hangzhou}
  \country{China}
}
\email{gaoangwang@intl.zju.edu.cn}


\begin{abstract}
Sports video analysis is crucial for athletic analytics and broadcasting enhancement. Dense sports video reasoning, however, demands a fine-grained understanding of numerous small-scale, highly interactive, and visually homogeneous entities (e.g., players sharing identical uniforms, the ball) across long temporal contexts. Current Large Multimodal Models (LMMs) inherently struggle with such dense visual complexities. Due to the lack of fine-grained visual details, these models often over-rely on textual priors to guess answers, especially when distinguishing visually similar actions and players. To address this, we propose \textbf{SportsGrounder}, a framework that leverages an open-vocabulary visual expert to aid interleaved grounding specifically for dense sports video reasoning. To achieve precise spatial localization, we extract domain-guided object proposals and introduce an Interleaved Grounding Fusion (IGF) mechanism. The IGF frame-by-frame integrates explicit bounding box coordinates and implicit visual semantics with global grid features. This design preserves strict temporal alignment and prevents sequence length explosion. Furthermore, we design an Action-Aware Supervision (AAS) module that directly regularizes the model's hidden states, forcing the network to learn accurate motion representations rather than relying on language bias. Optimized with Mixed Preference Optimization (MPO) to better distinguish deceptive distractors, our extensive experiments on newly curated dense sports VQA datasets (derived from SoccerNet and FineSports) demonstrate that SportsGrounder significantly improves fine-grained reasoning and achieves state-of-the-art accuracy.
\end{abstract}




\maketitle

\begin{figure}[t]
  \centering
  \includegraphics[width=\linewidth]{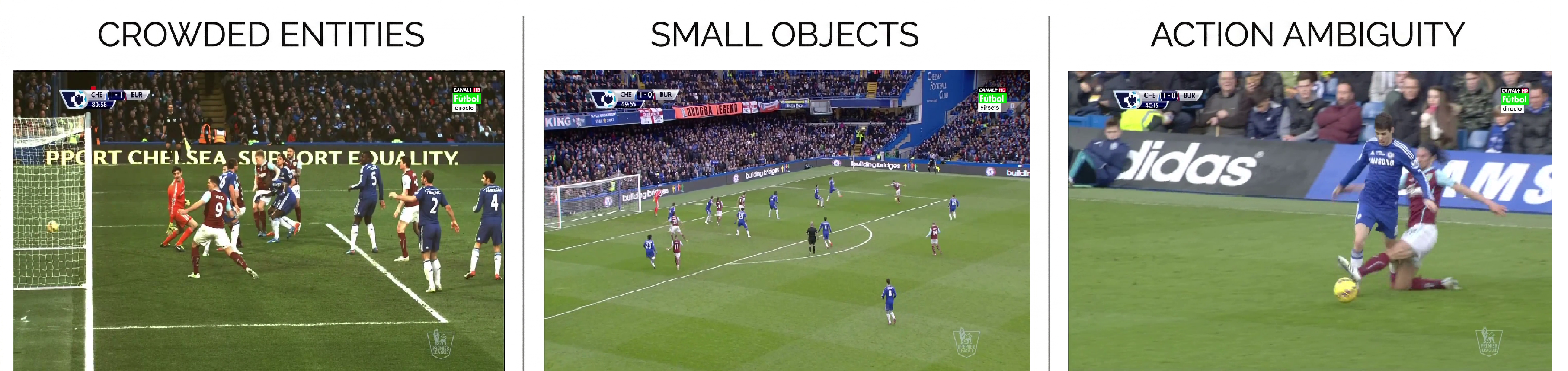}
  \caption{\textbf{Core challenges in fine-grained dense sports broadcast video grounding.} (Left) Crowded Entities: Multiple players with similar appearances in close proximity; (Middle) Small Objects: Crucial entities like the ball occupy very few pixels in broadcast views; (Right) Action Ambiguity: Visually similar initial states (e.g., a kicking motion) can lead to distinct fine-grained actions like passing or dribbling.}
\label{fig:teaser}
\end{figure}

\section{Introduction}

Sports video understanding is of great significance for athletic analytics, tactical coaching, and broadcasting enhancement, driving a major paradigm shift from coarse-grained event classification to highly granular, deductive video reasoning \cite{giancola2025soccernet, xia2026sportr, yang2026soccermaster, chen2025finequest}. In professional sports broadcasts, dense video reasoning requires a fine-grained understanding of numerous small-scale, visually similar entities (e.g., players of the same team, referees, the ball) and their rapid, complex interactions over extended temporal horizons. While the recent advancement of Large Multimodal Models (LMMs) has significantly improved general Video Question Answering (VQA) \cite{luo2025videoautoarena, ye2025rethinking, fu2025video}, directly applying existing LMMs to dense sports scenarios reveals critical limitations, as evidenced by their sub-optimal performance on emerging multi-agent and rule-grounded benchmarks \cite{liu2025f3set, salehi2024actionatlas, xia2025sportu, li2024sportsqa}.

Recently, Large Multimodal Models (LMMs) have demonstrated remarkable capabilities in general video understanding and open-ended visual question answering. By leveraging powerful Large Language Models (LLMs) alongside robust visual encoders, these architectures can seamlessly process interleaved vision-language inputs, enabling impressive zero-shot reasoning and rich semantic comprehension across diverse domains.

However, several challenges remain when applying these general-purpose LMMs to dense, highly dynamic scenarios such as professional sports broadcasts. (1) \textit{Severe loss of fine-grained spatial details.} Most contemporary LMMs rely on standard Vision Transformers (ViTs) to extract holistic grid features. The inherent patch-based processing and global pooling operations easily discard the subtle movements and complex spatial geometries of small, distant, and visually homogeneous entities (e.g., players sharing identical uniforms), which are crucial for accurate sports reasoning \cite{hong2025motionbench, deng2025motiongrounded}. Achieving pixel-level spatio-temporal alignment is paramount, yet existing architectures struggle to perceive micro-level motion dynamics without dedicated structural interventions \cite{munasinghe2025videoglamm, wang2024groundedvideollm, yang2025timeexpert, li2025unitime}. (2) \textit{Systemic over-reliance on language priors.} Because the extracted micro-level visual evidence is often blurry, occluded, or ambiguous, models tend to ignore the visual input and guess the answers based on dataset biases or textual cues \cite{park2025assessing, loginova2025addressing}. This modality bias is deeply problematic in the sports domain where actions share similar macro-level semantics; an LMM might hallucinate a ``shot'' instead of a ``pass'' simply due to linguistic likelihood, failing rudimentary counterfactual stress tests and exposing a brittle temporal comprehension \cite{howard2025uncovering, sethuraman2026stress, yang2026akdc, liao2025divide}. (3) \textit{Inefficient and destructive integration of visual grounding.} To provide precise localization and resolve this representation bottleneck, some methods introduce external object detectors \cite{tang2025adaptive, gong2025devil, xu2024finesports}. Yet, naively appending hundreds of object tokens or textual bounding boxes at the end of the global visual sequence destroys the inherent temporal structure of the video \cite{wang2026scener1, wang2025videollamb}. Moreover, this concatenation quadratically increases the sequence length, leading to an unacceptable self-attention computational overhead that cripples inference speeds in long-context scenarios \cite{yang2025thinking, zhou2026llava4d, nguyen2025hyperglm}.

To directly address these challenges, we propose \textbf{SportsGrounder}, to the best of our knowledge, the first framework that leverages proposal-aided interleaved grounding for dense sports video reasoning. Recognizing the inefficiency of using all object proposals in a crowded broadcast frame, we utilize an open-vocabulary visual expert (OV-DINO \cite{wang2024ovdino}) combined with a domain-specific vocabulary to select only the top-$K$ most relevant entities. To fuse these object-level details without destroying the temporal sequence, we propose the Interleaved Grounding Fusion (IGF) mechanism. Instead of globally concatenating tokens, IGF integrates explicit bounding box coordinates and implicit semantic features with the global grid features frame by frame, akin to recent advancements in grounded chain-of-thought processing \cite{linghu2026scenecot}. This simple yet effective design preserves temporal alignment and successfully avoids sequence length explosion. To ensure the model focuses on actual visual movements rather than exploiting language bias, we introduce an Action-Aware Supervision (AAS) module. Drawing inspiration from recent verifiable action regularizers \cite{yang2026guilibra, rao2025multiagent}, AAS provides explicit supervision on the terminal hidden states of the LMM during action-related queries, forcing the visual projectors to learn accurate, bias-resistant motion representations. Finally, the entire model is trained with a Mixed Preference Optimization (MPO) strategy. By optimizing over sets of preferred and dispreferred responses, MPO refines the decision boundary to successfully distinguish the hard-negative options commonly found in sports VQA, overcoming the inherent limitations of standard pairwise preference learning \cite{gupta2026multi, truong2026phidpo, zhang2025direct, huang2025longvpo, wang2024mpo}.

In summary, our main contributions are as follows:
\begin{itemize}
    \item We introduce SportsGrounder, the first proposal-aided reasoning framework specifically designed for dense sports VQA, which successfully mitigates the over-reliance on language priors by integrating an open-vocabulary visual expert.
    \item We propose the Interleaved Grounding Fusion (IGF) mechanism, which frame-by-frame combines explicit spatial anchors with global contexts, maintaining temporal alignment while keeping the sequence length highly efficient.
    \item We design an Action-Aware Supervision (AAS) module and utilize the MPO strategy to explicitly regularize the representation space, driving the model to focus on accurate motion dynamics against deceptive distractors.
    \item Extensive experiments on our newly curated dense sports VQA datasets (based on SoccerNet and FineSports) demonstrate that SportsGrounder achieves state-of-the-art performance and significantly improves fine-grained reasoning capabilities.
\end{itemize}

\section{Related Work}
\subsection{Large Multimodal Models for Dense Video Reasoning}
Recent Large Multimodal Models (LMMs) have made significant strides in general video understanding~\cite{fu2025video, zohar2025apollo, feng2025videor1, li2025f16, wang2025internvl35}, yet dense scenes with visually similar, small-scale entities expose a persistent failure mode: without reliable visual evidence, models tend to bypass genuine visual reasoning and instead exploit statistical language priors to guess answers~\cite{tong2024cambrian}. The Video-MME benchmark~\cite{fu2025video} systematically characterizes this gap between coarse-grained and fine-grained performance. SlowFocus~\cite{nie2024slowfocus} partially addresses it by identifying query-relevant segments and performing dense local sampling, while LLaVA-ST~\cite{li2025llavast} introduces Language-Aligned Positional Embedding and a Spatial-Temporal Packer for joint spatial and temporal localization. MotionBench~\cite{hong2025motionbench} further reveals that state-of-the-art VLMs fall below 60\% accuracy on fine-grained motion perception tasks, motivating Through-Encoder fusion for motion-level understanding. Grounded-VideoLLM~\cite{wang2024groundedvideollm} sharpens moment localization by introducing discrete temporal tokens and a dedicated temporal stream. Despite these advances, none of these works target the specific failure mode of dense sports video, where the language bias problem is especially acute due to the high visual similarity between action categories and players. This limitation directly motivates our Action-Aware Supervision and object-centric branch.

\subsection{Open-Vocabulary Detection and Grounding for Video LMMs}
Integrating explicit spatial grounding into LMMs requires both a capable open-vocabulary detector and an efficient fusion strategy. OV-DINO introduces language-aware selective fusion for accurate zero-shot detection~\cite{wang2024ovdino}, while Video-GroundingDINO generates continuous spatio-temporal tubes for open-vocabulary video grounding~\cite{wasim2024videogrounding}. At finer granularities, RGA3 accepts arbitrary visual prompts via Spatial-Temporal Overlay Modules~\cite{wang2025rga3}, and VideoGLaMM achieves pixel-level grounded video conversations~\cite{munasinghe2025videoglamm}. For action-centric understanding, ActionVOS segments state-changing objects using action-guided focal loss~\cite{ouyang2024actionvos}, and Kwon et al.\ refine this via question-conditioned keyframe selection~\cite{kwon2026learning}. Beyond Bare Queries constructs on-the-fly 3D scene graphs for multi-hop deductive reasoning~\cite{linok2025beyond}. On the preference optimization side, DPO~\cite{rafailov2023dpo} establishes the foundational framework for aligning model outputs with relative preference signals, and the MPO framework~\cite{wang2024mpo} extends this to multimodal reasoning by jointly optimizing preference, quality, and generative objectives. Our framework draws on these two lines of work, combining an OV-DINO-based grounding module with MPO training to calibrate the model's decision boundary against hard-negative distractors common in sports VQA.

\subsection{Sports Video Analytics}
Sports video analytics provides a demanding evaluation setting for multi-agent tracking and rule-based logical deduction. The SoccerNet benchmark suite~\cite{cioppa2022scaling} has driven progress in complex tasks such as game state reconstruction~\cite{giancola2025soccernet}, and X-VARS introduces explainability into sports officiating by aligning natural language rules with multi-view visual evidence~\cite{held2024xvars}. Foundation models like SoccerMaster~\cite{yang2026soccermaster} have emerged to unify diverse soccer understanding tasks, while SoccerBench assesses knowledge-driven multi-agent reasoning across comprehensive soccer scenarios~\cite{rao2025multiagent}. SportR further mandates explicit visual grounding by requiring models to output precise bounding boxes for rule infractions~\cite{xia2026sportr}. For fine-grained perception, FineSports provides hierarchical, multi-person annotations to evaluate models on high-speed, obstructed interactions~\cite{xu2024finesports}. However, existing benchmarks either lack systematic evaluation of language bias or do not assess fine-grained action disambiguation under dense entity conditions. In this work, we primarily evaluate on SoccerNet and conduct generalization experiments on FineSports to validate the robustness of our framework.

\section{Methodology}

\begin{figure*}[t]
  \centering
  \includegraphics[width=\linewidth]{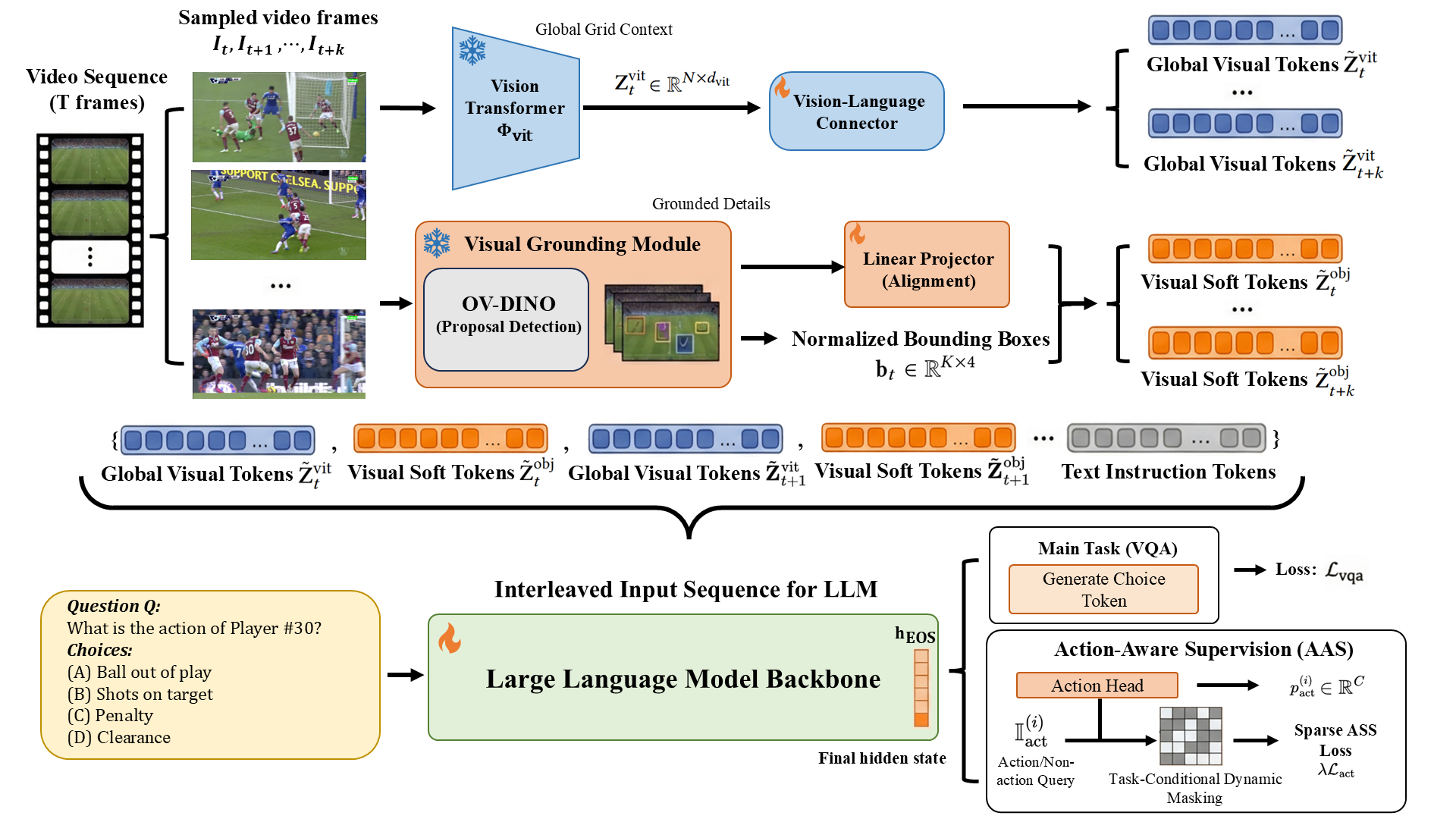}
  \caption{The overall architecture of SportsGrounder. Given a dense sports video and a text prompt, the framework first employs a proposal-aided representation strategy. A Vision Transformer captures the global grid context, while a pre-trained open-vocabulary detector (OV-DINO) leverages a text encoder to explicitly select and ground key domain-specific entities. These global and local representations are then chronologically combined frame-by-frame via the Interleaved Grounding Fusion (IGF) mechanism to preserve strict temporal correspondence. The fused visual tokens and text instructions are processed by the InternVL3.5 backbone. From the final hidden state, the model generates the final VQA response and simultaneously routes features to an auxiliary Action-Aware Supervision (AAS) head for dynamic motion regularization.}
\label{fig:model}
\end{figure*}


To address the inherent challenge of fine-grained spatio-temporal reasoning in dense sports scenarios, we introduce \textbf{SportsGrounder}, a Large Multimodal Model (LMM). As illustrated in Figure \ref{fig:model}, our architecture seamlessly integrates an open-vocabulary visual expert with a powerful LMM backbone to achieve precise visual grounding without overwhelming the sequence length. The overall framework consists of three core components: (1) Proposal-Aided Representation; (2) Interleaved Grounding Fusion (IGF); and (3) Action-Aware Supervision (AAS). To ensure optimal adaptation and prevent catastrophic forgetting, SportsGrounder is trained via a progressive curriculum comprising representation alignment, parameter-efficient fine-tuning (PEFT), and Mixed Preference Optimization (MPO).

\subsection{Proposal-Aided Representation}
Given a video instance $\mathcal{V}$, we uniformly sample $T$ frames, denoted as $\{I_1, I_2, \dots, I_T\}$. To capture a comprehensive understanding of the dynamic scene, we propose a proposal-aided representation strategy that disentangles the holistic background context from the fine-grained foreground entities.

\paragraph{Global Grid Context.}
For the global branch, we utilize a pre-trained Vision Transformer $\Phi_{\text{vit}}$ \cite{wang2025internvl35} to extract the dense grid features. By removing the global pooling layer, we preserve the dense grid representations. For the $t$-th frame $I_t$, the global feature map is obtained as:
\begin{equation}
    \mathbf{Z}_t^{\text{vit}} = \Phi_{\text{vit}}(I_t) \in \mathbb{R}^{N \times d_{\text{vit}}}
\end{equation}
where $N$ is the number of patch tokens and $d_{\text{vit}}$ is the hidden dimension of the ViT. This branch ensures the model retains a holistic understanding of the field layout, crowd context, and overall scene structure.

\paragraph{Language-Aware Object Grounding.}
To explicitly capture fine-grained entities, we introduce an object-centric branch powered by OV-DINO \cite{wang2024ovdino}, a state-of-the-art open-vocabulary detector equipped with a Language-Aware Selective Fusion (LASF) module. In its native configuration, the model decodes a dense set of selective fusion queries $\mathbf{Q}_{\text{sf}} \in \mathbb{R}^{M \times d_{\text{dino}}}$, where $M=900$ and $d_{\text{dino}}$ is the hidden dimension. However, directly interleaving $M$ queries per frame would computationally overwhelm the LMM's context window.

To distill critical visual semantics, we propose a Domain-Guided Top-$K$ Selection mechanism. We define a sport-specific semantic vocabulary $\mathcal{V}_{\text{sports}}$ and map it into text embeddings $\mathbf{E}_t \in \mathbb{R}^{C \times d_{\text{dino}}}$, where $C$ is the number of target categories. Following the alignment mechanics of OV-DINO, we compute the cross-modality similarity score matrix $\mathbf{S}$:
\begin{equation}
    \mathbf{S} = \mathcal{F}_c(\mathbf{Q}_{\text{sf}}) \otimes \mathbf{E}_t^\top \in \mathbb{R}^{M \times C}
\end{equation}
where $\mathcal{F}_c$ is the linear projection for class embeddings, and $\otimes$ denotes standard matrix multiplication. By evaluating the maximum confidence scores across the $\mathcal{V}_{\text{sports}}$ categories in $\mathbf{S}$, we filter the $M$ proposals and strictly retain only the top-$K$ queries ($K \ll M$) that exhibit the highest alignment with our core sports elements. The filtered semantic object features for frame $I_t$ are concisely represented as $\mathbf{Z}_t^{\text{sem}} \in \mathbb{R}^{K \times d_{\text{dino}}}$.

Simultaneously, recognizing that dense sports reasoning necessitates rigorous geometric precision, we extract the explicit normalized bounding box coordinates $\mathbf{b}_t \in \mathbb{R}^{K \times 4}$ associated with these selected queries. This dual extraction yields a hybrid representation for each entity, comprising both implicit high-dimensional vectors and explicit spatial anchors.

\subsection{Interleaved Grounding Fusion (IGF)}
To map the disparate visual representations into the unified latent space of the large language model ($d_{\text{llm}}$), we employ two distinct MLP projectors, $\mathcal{P}_{\text{vit}}$ and $\mathcal{P}_{\text{dino}}$. The global grid context is projected as:
\begin{equation}
    \tilde{\mathbf{Z}}_t^{\text{vit}} = \mathcal{P}_{\text{vit}}(\mathbf{Z}_t^{\text{vit}}) \in \mathbb{R}^{N \times d_{\text{llm}}}
\end{equation}

For the object-centric branch, we construct a \textit{Hybrid Entity Token} sequence. Specifically, for the $k$-th selected entity in frame $I_t$, we first tokenize its explicit bounding box $\mathbf{b}_{t,k} = [x_1, y_1, x_2, y_2]$ and map it through the LLM's text embedding layer $\Phi_{\text{emb}}$ to obtain a discrete textual embedding $\mathbf{E}_{t,k}^{\text{box}} \in \mathbb{R}^{L_{\text{box}} \times d_{\text{llm}}}$, where $L_{\text{box}}$ denotes the average token length of the coordinate string. 

Subsequently, we project its corresponding semantic visual embedding $\mathbf{Z}_{t,k}^{\text{sem}}$ (a $1 \times d_{\text{dino}}$ slice from $\mathbf{Z}_t^{\text{sem}}$) via $\mathcal{P}_{\text{dino}}$, yielding a spatial-aware semantic vector $\mathcal{P}_{\text{dino}}(\mathbf{Z}_{t,k}^{\text{sem}}) \in \mathbb{R}^{1 \times d_{\text{llm}}}$. By concatenating the explicit coordinate tokens with the implicit visual vector, the joint object representation for all $K$ entities in frame $I_t$ is formulated as:
\begin{equation}
    \tilde{\mathbf{Z}}_t^{\text{obj}} = \bigoplus_{k=1}^{K} \left( \mathbf{E}_{t,k}^{\text{box}} \oplus \mathcal{P}_{\text{dino}}(\mathbf{Z}_{t,k}^{\text{sem}}) \right) \in \mathbb{R}^{K(L_{\text{box}} + 1) \times d_{\text{llm}}}
\end{equation}
where $\oplus$ denotes concatenation along the token sequence dimension, and $\bigoplus_{k=1}^{K}$ indicates the sequential concatenation of these hybrid representations across all $K$ selected objects.

Traditional methods tend to append all tokens at the end of the sequence, but this disrupts temporal correspondence and weakens local cross-attention. To fix this, we propose Interleaved Grounding Fusion (IGF), which chronologically interleaves global contexts and hybrid object tokens on a frame-by-frame basis:
\begin{equation}
    \mathbf{H}_{\text{vis}} = \left[ \tilde{\mathbf{Z}}_1^{\text{vit}} \oplus \tilde{\mathbf{Z}}_1^{\text{obj}} \oplus \dots \oplus \tilde{\mathbf{Z}}_T^{\text{vit}} \oplus \tilde{\mathbf{Z}}_T^{\text{obj}} \right] \in \mathbb{R}^{T(N + K(L_{\text{box}} + 1)) \times d_{\text{llm}}}
\end{equation}


By drastically distilling the redundant object queries via the vocabulary-guided selection ($K \ll 900$) and mapping them into a compact hybrid representation, the resultant sequence length is significantly constrained. This interleaved topology effectively circumvents the quadratic explosion of the self-attention matrix associated with naive fusion $\mathcal{O}((T(N+K\cdot L_{\text{box}}))^2)$, ensuring that the LMM simultaneously accesses the macro-environment and geometrically-anchored micro-entities at each timestamp with profound computational efficiency.

\subsection{Action-Aware Supervision (AAS) Mechanism}
Optimizing the LMM solely via standard auto-regressive text generation loss often leads the model to over-rely on language priors, neglecting the complex motion dynamics present in dense sports videos. To enforce rigorous spatio-temporal reasoning and ensure the model grounds its predictions in actual visual evidence, we introduce an auxiliary Action-Aware Supervision (AAS) branch.

Let $\mathbf{H}_{\text{text}}$ be the token embeddings of the prompt instruction (containing $\mathcal{Q}$ and $\mathcal{C}$). The complete input sequence to the LMM is $\mathbf{H}_{in} = [\mathbf{H}_{\text{vis}} \oplus \mathbf{H}_{\text{text}}]$. After the causal forward pass, we obtain the sequence of hidden states $\mathbf{E} \in \mathbb{R}^{L \times d_{\text{llm}}}$. 

We extract the final hidden state $\mathbf{h}_{\text{EOS}}$ corresponding to the end-of-sequence token. Given that standard attention mechanisms propagate prior context forward, $\mathbf{h}_{\text{EOS}}$ encapsulates the global semantic aggregation of the entire video-text input. We map this state to the action semantic space to predict the discrete action category $\hat{y}_{\text{act}}$:
\begin{equation}
    p_{\text{act}}^{(i)} = \text{Softmax}(\mathbf{W}_{\text{act}} \mathbf{h}_{\text{EOS}}^{(i)} + \mathbf{b}_{\text{act}}) \in \mathbb{R}^{C}
\end{equation}
where $i$ denotes the sample index within a batch, $\mathbf{W}_{\text{act}} \in \mathbb{R}^{C \times d_{\text{llm}}}$ is the action projection head, and $C$ is the number of predefined action classes.

While causal attention mechanisms can naturally heavily weigh the adjacent textual inputs at the terminal tokens, the explicit backpropagation of $\mathcal{L}_{\text{act}}$ directly through $\mathbf{h}_{\text{EOS}}$ mitigates this tendency. This direct gradient flow compels the terminal state to actively aggregate vital spatio-temporal dynamics from the preceding visual sequence $\mathbf{H}_{\text{vis}}$, preventing the model from relying solely on the textual context.

\paragraph{Task-Conditional Dynamic Masking.}
Dense sports VQA datasets inherently comprise a heterogeneous mixture of tasks, encompassing not only action recognition but also spatial reasoning and object localization. Enforcing action supervision on non-action queries would inevitably lead to negative transfer and distort spatial feature representations. To circumvent this, we propose a task-conditional dynamic masking mechanism. We define an indicator function $\mathbb{I}_{\text{act}}^{(i)} \in \{0, 1\}$ derived \textit{a priori} from the dataset's meta-annotations, where $\mathbb{I}_{\text{act}}^{(i)} = 1$ if the ground-truth task category of the $i$-th query pertains to action semantics, and $0$ otherwise. The masked AAS loss over a mini-batch of size $B$ is formulated as:
\begin{equation}
    \mathcal{L}_{\text{act}} = - \frac{1}{\sum_{i=1}^{B} \mathbb{I}_{\text{act}}^{(i)} + \epsilon} \sum_{i=1}^{B} \mathbb{I}_{\text{act}}^{(i)} \sum_{c=1}^{C} y_{\text{act},c}^{(i)} \log(p_{\text{act},c}^{(i)})
\end{equation}
where $y_{\text{act}}^{(i)}$ is the ground-truth action label for the $i$-th sample, and $\epsilon$ is a small smoothing constant to prevent zero-division. To ensure numerical stability and consistent gradient flow during training, we implement a balanced task-sampling strategy during mini-batch construction, ensuring that each batch contains a minimum proportion of action-related queries where $\sum \mathbb{I}_{\text{act}}^{(i)}>0$.

By employing this sparse supervisory signal, the LMM acts as a task-aware router: it activates the fine-grained action classification head exclusively when motion dynamics are interrogated, while preserving unbiased spatio-temporal representations for pure localization tasks.

\subsection{Three-Stage Training Curriculum}
To ensure stable convergence and optimal feature alignment, the framework is optimized via a progressive three-stage curriculum.

\paragraph{Stage 1: Asymmetric Representation Alignment.}
To prevent catastrophic forgetting of the pre-trained generalized visual capabilities, we adopt an asymmetric freezing strategy. We freeze the visual encoders ($\Phi_{\text{vit}}$, $\Phi_{\text{dino}}$), the large language model, and the pre-trained global connector $\mathcal{P}_{\text{vit}}$. We exclusively optimize the parameters of the newly initialized object projector, denoted as $\theta_{\text{dino}}$, using image-text pairs. The objective is to minimize the auto-regressive negative log-likelihood of the ground-truth text tokens $\mathcal{Y} = \{y_1, y_2, \dots, y_L\}$:
\begin{equation}
    \min_{\theta_{\text{dino}}} \mathcal{L}_{\text{align}} = - \sum_{j=1}^{L} \log P(y_j \mid \mathbf{H}_{\text{vis}}(\theta_{\text{dino}}), y_{<j})
\end{equation}
where $\mathbf{H}_{\text{vis}}(\theta_{\text{dino}})$ explicitly denotes that the interleaved visual sequence is dynamically updated solely through the gradients backpropagated to $\mathcal{P}_{\text{dino}}$. This strict partial optimization ensures that the nascent object queries seamlessly align towards the established manifold of the LLM without perturbing the robust grid representations.

\paragraph{Stage 2: Parameter-Efficient Supervised Fine-Tuning (PEFT)}
Following the initial alignment, performing full-parameter fine-tuning on the massive LMM backbone is computationally prohibitive and prone to catastrophic forgetting. Therefore, we freeze the dense pre-trained weights of the LMM ($\Phi_{\text{llm}}$) and employ Low-Rank Adaptation (LoRA). Specifically, for the pre-trained weight matrices $\mathbf{W}_0$, we inject trainable low-rank decomposition matrices such that the forward pass becomes $\mathbf{W} = \mathbf{W}_0 + \mathbf{B}\mathbf{A}$, where $\mathbf{B}$ and $\mathbf{A}$ contain the LoRA parameters $\theta_{\text{LoRA}}$. 

Concurrently, both the global and object visual projectors ($\mathcal{P}_{\text{vit}}$ and $\mathcal{P}_{\text{dino}}$) are fully unfrozen to undergo end-to-end task-specific optimization. Thus, the active learnable parameter set is strictly confined to $\Theta_{\text{SFT}} = \{\theta_{\text{LoRA}}, \theta_{\mathcal{P}_{\text{vit}}}, \theta_{\mathcal{P}_{\text{dino}}}\}$. Given the interleaved multimodal context prefix $\mathbf{X} = [\mathbf{H}_{\text{vis}} \oplus \mathbf{H}_{\text{text}}]$ and the target answer sequence $\mathcal{Y}_{\text{ans}} = \{y_1, \dots, y_{L_{\text{ans}}}\}$, the auto-regressive generation loss is formulated as:
\begin{equation}
    \mathcal{L}_{\text{vqa}} = - \sum_{t=1}^{L_{\text{ans}}} \log P_{\Theta_{\text{SFT}}}(y_t \mid \mathbf{X}, y_{<t})
\end{equation}
The joint objective seamlessly integrates our proposed masked AAS mechanism to regularize the motion representation:
\begin{equation}
    \mathcal{L}_{\text{SFT}} = \mathcal{L}_{\text{vqa}} + \lambda \mathcal{L}_{\text{act}}
\end{equation}

\paragraph{Stage 3: Mixed Preference Optimization (MPO)}
To rigidly calibrate the decision boundary against visually deceptive distractors in sports scenarios, we leverage the Mixed Preference Optimization (MPO) framework natively introduced by our base model, InternVL3.5 \cite{wang2025internvl35}. Rather than formulating a novel alignment paradigm, we adopt this established protocol to ensure that the LMM comprehensively learns the relative preference, absolute response quality, and fundamental generative structure simultaneously.

For a given context $\mathbf{X}$, we construct preference pairs comprising the ground-truth reasoning process $y_c \in \mathcal{Y}_p$ (positive set) and a hard-negative response $y_r \in \mathcal{Y}_n$ (negative set). Following the official InternVL3.5 specification, the total optimization objective is a weighted combination:
\begin{equation}
    \mathcal{L}_{\text{MPO}} = \mathcal{L}_{p} + \alpha \mathcal{L}_{q} + \beta \mathcal{L}_{g}
\end{equation}
Specifically, $\mathcal{L}_{p}$ acts as the preference loss (e.g., DPO) which strictly updates the active LoRA parameters $\theta_{LoRA}$ to optimize the relative margin:
\begin{equation}
    \mathcal{L}_{p} = - \log \sigma \left( \tau \log \frac{\pi_{\theta_{\text{LoRA}}}(y_c \mid \mathbf{X})}{\pi_{\text{ref}}(y_c \mid \mathbf{X})} - \tau \log \frac{\pi_{\theta_{\text{LoRA}}}(y_r \mid \mathbf{X})}{\pi_{\text{ref}}(y_r \mid \mathbf{X})} \right)
\end{equation}
where $\pi_{\text{ref}}$ is the frozen reference model initialized directly from the Stage 2 checkpoint. Concurrently, $\mathcal{L}_{q}$ explicitly evaluates the absolute quality of the response, and $\mathcal{L}_{g}$ enforces auto-regressive generation constraints on the positive samples to prevent format collapse during alignment.

\section{Experiments}

\begin{figure}[t]
    \centering
    \includegraphics[width=0.8\linewidth]{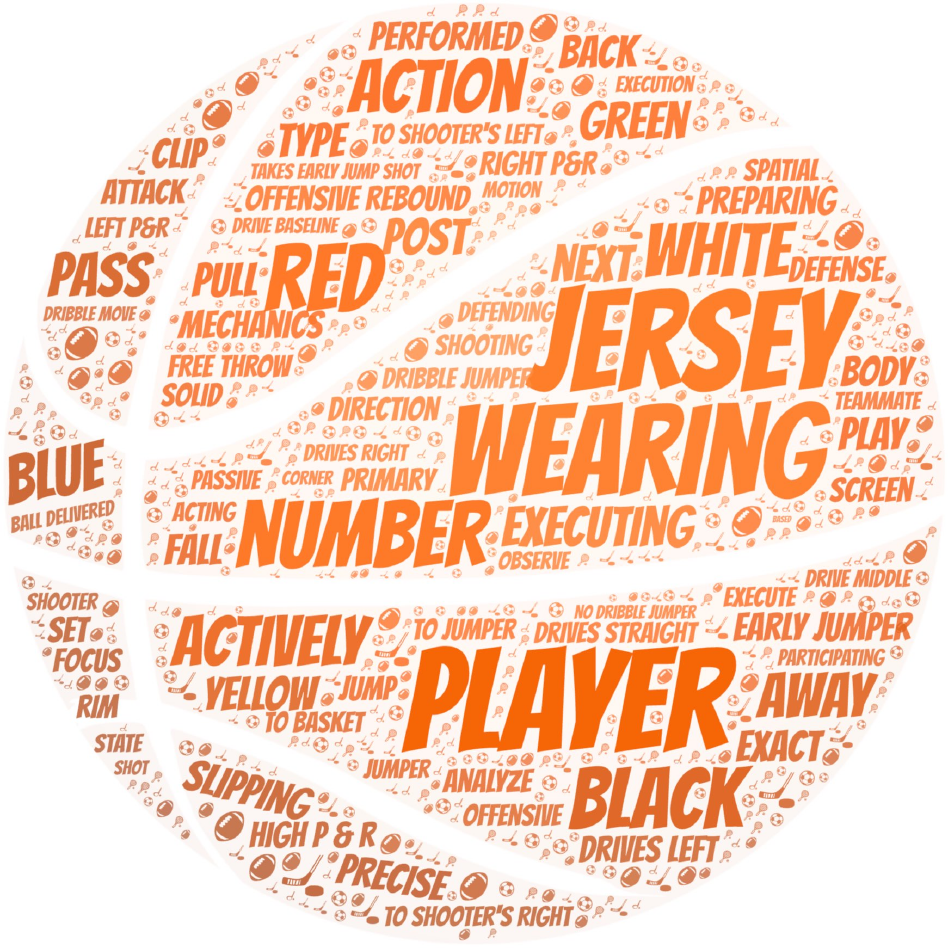}
    \caption{Word cloud visualization of the generated question-answer pairs from the FineSports dataset, highlighting the diverse tactical and action-oriented vocabulary required for dense sports reasoning.}
    \label{fig:wordcloud}
\end{figure}
\subsection{Experimental Settings}

\begin{figure*}[t]
    \centering
    \includegraphics[width=\linewidth]{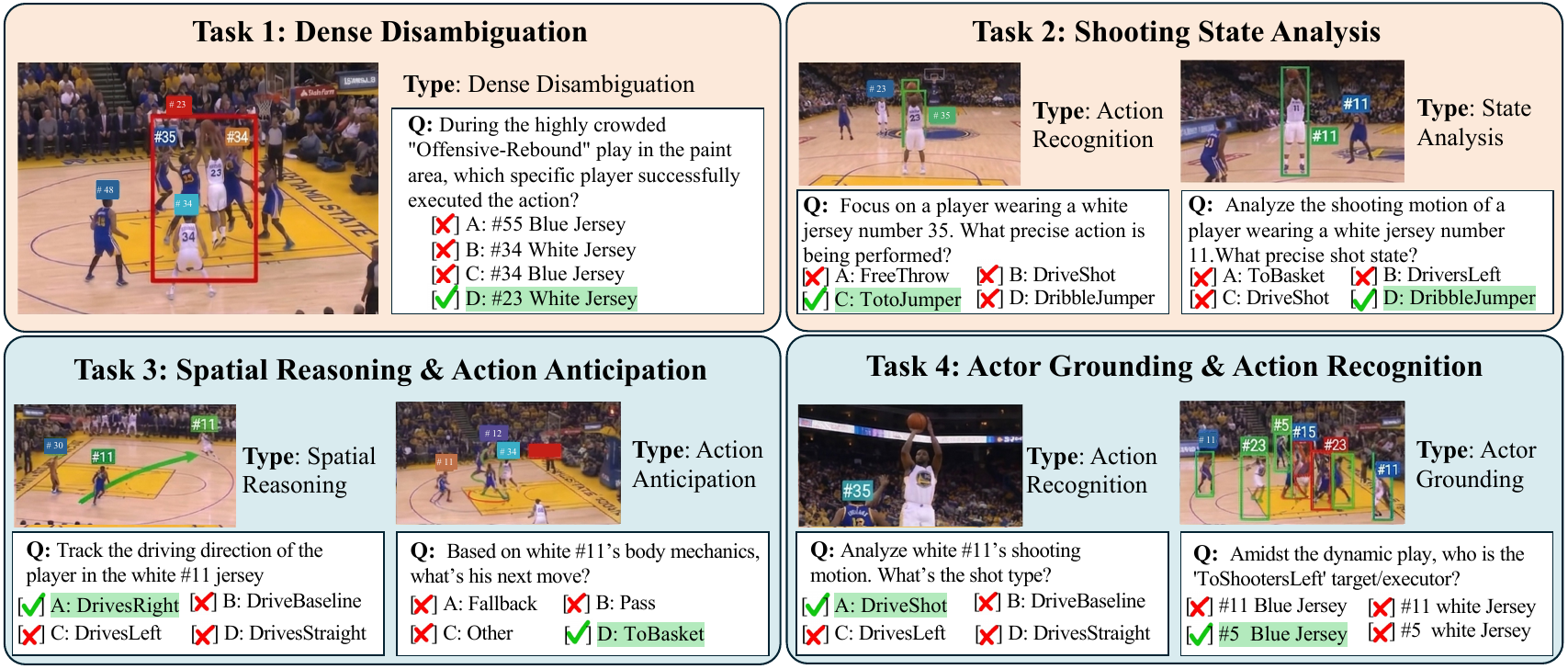}
    \caption{Qualitative examples from our curated FineSports QA dataset. The benchmark encompasses diverse fine-grained tasks, such as dense disambiguation, shooting state analysis, spatial reasoning, and actor grounding, requiring the model to possess robust spatio-temporal comprehension and accurate entity localization in highly cluttered scenes.}
    \label{fig:case_study}
\end{figure*}

\paragraph{Datasets and Annotations.}
To rigorously evaluate our proposed framework, we conduct extensive experiments on two distinct, densely annotated sports video datasets: SoccerNet (soccer)~\cite{cioppa2022scaling} and FineSports (basketball)~\cite{xu2024finesports}. For SoccerNet, we clip short segments around annotated timestamps and uniformly sample sparse keyframes; for FineSports, we directly use the pre-extracted complete sequences (containing up to 8 frames per clip) to maximally preserve motion dynamics. Standard Video QA benchmarks often lack the fine-grained interactions present in professional sports. Therefore, we specifically selected these two datasets because they provide comprehensive multi-modal annotations, including precise action labels, spatial bounding boxes, and fine-grained entity attributes (e.g., player jersey numbers). Crucially, the action labels serve a dual purpose: they act as the foundation for constructing our action recognition QA pairs and define the ground-truth label space for our Action-Aware Supervision (AAS) module. 
For \textbf{SoccerNet}, which encompasses 17 distinct action categories (e.g., \textit{Goal}, \textit{Substitution}, \textit{Penalty}), we generated a total of 26k QA pairs, split into 25k for training and 1k for testing. Conversely, \textbf{FineSports} features 24 basketball-specific action labels (e.g., \textit{DribbleMove}, \textit{NoDribbleJumper}, \textit{Post-Up}). Based on these, we generated 24k QA pairs (23k for training and 1k for testing). Due to the fundamental divergence in action taxonomies and visual scenes, the models for SoccerNet and FineSports are trained independently. 

\paragraph{QA Construction Strategy.}
Leveraging the rich metadata from both datasets, we programmatically construct diverse 4-option multiple-choice QA pairs designed to holistically probe dense sports reasoning. Rather than isolating distinct tasks, our generation strategy seamlessly integrates spatial comprehension and action recognition. For \textbf{FineSports}, the questions evaluate multifaceted capabilities such as actor grounding, dense disambiguation, shooting state analysis, and interaction role identification. Figure \ref{fig:case_study} provides concrete qualitative examples of these sub-tasks, illustrating how the model must accurately distinguish subtle action states amidst heavy occlusion. Similarly, for \textbf{SoccerNet}, the generation engine covers team comparison, player identity, topological location, and action classification. To ensure linguistic diversity and robust grounding, the prompts utilize varied referring expressions to localize the target subject, such as referencing a player's jersey number, relative on-field position, or distinctive jersey color. Crucially, to prevent the LMM from exploiting textual shortcuts, we strictly exclude any explicit $(x,y)$ bounding box coordinates from the questions in our evaluation sets. This constraint compels the model to genuinely align visual entities with the text rather than trivially matching spatial strings. To illustrate the overall linguistic complexity of our evaluation benchmark, Figure \ref{fig:wordcloud} visualizes the word cloud distribution of the generated QA pairs from the FineSports dataset.

\paragraph{Implementation Details.}
We instantiate \textbf{SportsGrounder} using InternVL3.5-2B~\cite{wang2025internvl35} as our base Large Multimodal Model. For the visual representation extraction, the global branch employs the default ViT encoder, while the object-centric local branch utilizes a pre-trained OV-DINO to extract top-$K$ semantic object tokens. For the object-centric local branch, we empirically set the top-$K$ selection hyperparameter to $K=15$. This value strikes an optimal balance between exhaustively capturing all active players in dense clusters and mitigating the computational overhead caused by irrelevant background noise (e.g., spectators). Crucially, both visual encoders are kept strictly frozen throughout the entire training process to preserve their robust foundational representations. Our progressive curriculum is efficiently structured into three stages. In Stage 1, since the default ViT connector $\mathcal{P}_{\text{vit}}$ is already well-aligned in InternVL3.5, we keep it frozen alongside the LLM. We solely initialize and pre-train the newly introduced object-centric projector ($\mathcal{P}_{\text{dino}}$, implemented as a two-layer MLP) for 1 epoch to align the OV-DINO features with the LLM's embedding space. Subsequently, the model undergoes 3 epochs of joint Supervised Fine-Tuning (SFT) in Stage 2, followed by 1 epoch of Mixed Preference Optimization (MPO) in Stage 3. During Stage 2 and 3 training, we preserve the pre-trained weights of the LLM backbone and apply Low-Rank Adaptation (LoRA) for parameter-efficient optimization, while fully updating both visual projectors ($\mathcal{P}_{\text{vit}}$ and $\mathcal{P}_{\text{dino}}$). For the multi-task loss formulation, we empirically set the action supervision weight $\lambda = 0.1$ to prevent the auxiliary objective from dominating the primary text generation. In the MPO phase, we strictly adhere to the default preference optimization recipe provided by InternVL3.5. Specifically, we set the DPO divergence margin $\beta=0.1$ and optimize all parameters using the AdamW optimizer coupled with a cosine learning rate decay schedule.

\paragraph{Baselines and Evaluation Protocol.}
The evaluation metric across all experiments is the top-1 accuracy (\%) on the 4-choice QA tasks. To ensure a rigorous and equitable comparison, we benchmark SportsGrounder against recent state-of-the-art VLMs under two settings: (1) \textit{Parameter-Efficient SFT}, where baselines are fine-tuned directly on our sports datasets using their default global visual inputs; and (2) \textit{Prompt Injection}, a stronger baseline where we explicitly format the OV-DINO bounding box coordinates as textual prompts and inject them into the baselines' input context. This allows us to accurately isolate the architectural superiority of our Interleaved Grounding Fusion (IGF) mechanism against merely providing spatial hints.

\subsection{Main Results}

Table \ref{tab:main_results} presents the quantitative comparison of SportsGrounder against various recent state-of-the-art VLMs on the SoccerNet and FineSports datasets. To provide a comprehensive evaluation, we decompose the overall reasoning performance into fine-grained sub-tasks based on their unique cognitive demands. To ensure a strictly fair evaluation, our SportsGrounder is built upon the InternVL3.5-2B backbone \cite{wang2025internvl35}. Similarly, all competitive baselines, including Qwen3-VL-2B-Instruct \cite{bai2025qwen3}, VideoLLaMA3-2B \cite{zhang2025videollama}, and the larger MiniCPM-V 4.0 \cite{yao2024minicpm}, are fine-tuned on our curated datasets using Low-Rank Adaptation (LoRA) rather than full-parameter fine-tuning.

\begin{table*}[t]
\centering
\caption{Main results on SoccerNet and FineSports datasets. All results are reported as Accuracy (\%) on 4-option multiple-choice questions. ``Size'' denotes the number of parameters of the LLM backbone. ``Prompt Injection'' denotes explicitly feeding bounding box coordinates into the LLM context as text. Best results are in \textbf{bold}.}
\label{tab:main_results}
\resizebox{\linewidth}{!}{
\begin{tabular}{l c ccccc ccccc}
\toprule
\multirow{2}{*}{\textbf{Method}} & \multirow{2}{*}{\textbf{Size}} & \multicolumn{5}{c}{\textbf{SoccerNet}} & \multicolumn{5}{c}{\textbf{FineSports}} \\
\cmidrule(lr){3-7} \cmidrule(lr){8-12}
& & \textbf{Overall} & Action & Team & Jersey & Spatial & \textbf{Overall} & Action & Ground. & Disamb. & Spatial \\
\midrule
\multicolumn{12}{l}{\textit{Standard SFT Baselines (w/ LoRA)}} \\
\midrule
VideoLLaMA3-2B       & 2B & 38.6 & 32.5 & 55.2 & 24.5 & 40.0 & 40.2 & 33.8 & 45.2 & 41.5 & 44.1 \\
Qwen3-VL-2B-Instruct & 2B & 42.3 & 36.4 & 59.8 & 29.3 & 43.0 & 43.5 & 37.9 & 48.6 & 44.2 & 47.6 \\
InternVL3.5-2B       & 2B & 43.7 & 37.8 & 61.4 & 31.1 & 43.8 & 44.8 & 38.5 & 49.9 & 45.5 & 48.3 \\
MiniCPM-V 4.0        & 4B & 47.5 & 41.2 & 64.1 & 37.8 & 46.8 & 48.2 & 42.6 & 53.4 & 48.7 & 51.5 \\
\midrule
\multicolumn{12}{l}{\textit{Prompt Injection Baselines (w/ Expert BBoxes as Text)}} \\
\midrule
VideoLLaMA3-2B + BBox       & 2B & 39.4 & 32.1 & 57.6 & 26.2 & 41.1 & 41.6 & 33.3 & 49.5 & 43.8 & 45.3 \\
Qwen3-VL-2B-Instruct + BBox & 2B & 43.8 & 36.0 & 62.3 & 30.8 & 44.3 & 45.3 & 37.5 & 52.8 & 46.9 & 48.5 \\
InternVL3.5-2B + BBox       & 2B & 45.2 & 37.3 & 63.8 & 32.5 & 45.0 & 46.6 & 38.2 & 54.2 & 48.1 & 49.5 \\
MiniCPM-V 4.0 + BBox        & 4B & 49.1 & 40.8 & 66.5 & \textbf{40.3} & 48.2 & 50.4 & 42.1 & 57.8 & 51.5 & 53.0 \\
\midrule
\textbf{SportsGrounder (Ours)} & 2B & \textbf{51.8} & \textbf{42.5} & \textbf{71.2} & 33.4 & \textbf{54.2} & \textbf{53.6} & \textbf{43.2} & \textbf{59.2} & \textbf{53.1} & \textbf{54.2} \\
\bottomrule
\end{tabular}
}
\end{table*}

\begin{table*}[t]
\centering
\caption{Ablation study of SportsGrounder on the SoccerNet dataset. ``Baseline (IGF only)'' indicates the model utilizes the Interleaved Grounding Fusion but is evaluated after Stage 2 standard SFT without the auxiliary AAS loss. Best results are in \textbf{bold}.}
\label{tab:ablation}
\begin{tabular}{lcc ccccc}
\toprule
\multirow{2}{*}{\textbf{Model Variant}} & \multirow{2}{*}{\textbf{w/ AAS Loss}} & \multirow{2}{*}{\textbf{w/ MPO (Stage 3)}} & \multicolumn{5}{c}{\textbf{SoccerNet Acc. (\%)}} \\
\cmidrule(lr){4-8}
& & & \textbf{Overall} & Action & Team & Jersey & Spatial \\
\midrule
Baseline (IGF only) & \ding{55} & \ding{55} & 47.2 & 38.4 & 66.5 & 32.1 & 48.6 \\
+ Action Supervision & \ding{51} & \ding{55} & 48.9 & 40.8 & 68.2 & 32.3 & 49.5 \\
+ Preference Opt. & \ding{55} & \ding{51} & 49.6 & 39.1 & 69.4 & 32.8 & 53.0 \\
\midrule
\textbf{SportsGrounder (Full)} & \ding{51} & \ding{51} & \textbf{51.8} & \textbf{42.5} & \textbf{71.2} & \textbf{33.4} & \textbf{54.2} \\
\bottomrule
\end{tabular}
\end{table*}

\paragraph{Superiority over Standard SFT}
As shown in Table \ref{tab:main_results}, breaking down the \textit{Overall} performance reveals extreme capability variance across different cognitive dimensions. For tasks demanding relatively coarse visual discrimination, such as \textit{Team Comparison} (distinguishing contrasting jersey colors), most baselines perform decently (reaching over 60\% accuracy). Conversely, fine-grained action recognition exhibits severe degradation (hovering around 32\%-41\%) due to the domain gap of unseen athletic motions. Even more challenging is the \textit{Jersey Number} query on SoccerNet, where typical 2B models plunge below 33\% accuracy. This plunge is logical: spotting blurred numbers on distant, fast-moving players heavily relies on brute-force Optical Character Recognition (OCR) capacities, which scale strongly with parameter size. Consequently, the 4B-parameter MiniCPM-V naturally dominates the OCR-heavy \textit{Jersey} category. 

Despite this expected gap in OCR, SportsGrounder explicitly overcomes the fundamental spatial and temporal bottlenecks. By achieving \textbf{51.8\% Overall} accuracy on SoccerNet, our 2B-parameter framework comprehensively outperforms all baselines, including the 4B MiniCPM-V. This demonstrates that dedicated Interleaved Grounding Fusion (IGF) is a far more efficient solution for complex, relationship-heavy topologies (\textit{Spatial Reas.}, \textit{Dense Disambig.}) than sheer parameter scaling.

\paragraph{Effectiveness against Prompt Injection.}
While explicit bounding box coordinate injection naturally inflates straightforward localization tasks (e.g., \textit{Actor Grounding} surges significantly), it acts as a double-edged sword. In dense spatial scenarios like \textit{Dense Disambiguation}, pure numbers fail to capture visual topologies, yielding diminishing returns. Crucially, overloading the context window with text-based geometry consistently distracts the LLM's temporal attention, leading to a persistent drop in \textit{Action} prediction across all baselines. Our framework fundamentally avoids this text-based distraction by weaving spatial clues as native visual tokens, thereby establishing consistent state-of-the-art performance in both overall and sub-task accuracies.

\subsection{Ablation Studies}

To systematically validate the contribution of our core designs, we perform ablation studies on the SoccerNet dataset. We investigate the impact of the Action-Aware Supervision (AAS) module and the Mixed Preference Optimization (MPO) training stage. The results are summarized in Table \ref{tab:ablation}.

\paragraph{Impact of Action-Aware Supervision (AAS)}
Removing the AAS branch yields the Baseline variant, which optimizes the LMM solely via text generation. Benefiting from the IGF architecture, this baseline already secures a solid \textit{Overall} accuracy of 47.2\% (outperforming standard InternVL3.5). However, its \textit{Action} recognition remains suppressed at 38.4\%. Activating $\mathcal{L}_{\text{act}}$ compels the network's terminal state to learn distinct motion dynamics, which independently drives a targeted +2.4\% surge in \textit{Action} accuracy while exerting a mild positive regularization effect on the overall visual representation (Overall rises to 48.9\%).

\paragraph{Impact of the Progressive Training Curriculum.}
Evaluating the base model before Stage 3 reveals vulnerability when distinguishing visually similar distractors. The introduction of Stage 3 MPO explicitly calibrates the decision boundaries against hard-negative choices. As evidenced in Table \ref{tab:ablation}, preference optimization profoundly benefits relational tasks, significantly boosting \textit{Spatial Reasoning} (+4.4\% on the baseline) and \textit{Team Comparison}, pushing the \textit{Overall} accuracy up to 49.6\%. Integrating both AAS and MPO achieves compounding gains, propelling the fully equipped SportsGrounder to the optimal 51.8\% Overall performance.

\section{Conclusion}

In this paper, we introduced \textbf{SportsGrounder}, a novel Large Multimodal Model explicitly tailored for fine-grained spatio-temporal reasoning in dense sports video question-answering. To overcome the inherent limitations of holistic visual encoders in cluttered and highly dynamic scenes, we proposed a proposal-aided representation architecture that synergizes global grid contexts with explicit, object-centric features from a pre-trained open-vocabulary detector. By chronologically weaving these representations via the proposed Interleaved Grounding Fusion (IGF) mechanism, our framework preserves strict temporal alignment without succumbing to sequence length explosion. Furthermore, the introduction of the Action-Aware Supervision (AAS) module effectively mitigates language priors, compelling the LMM to ground its predictions in actual motion dynamics. Extensive experiments on the SoccerNet and FineSports benchmarks demonstrate that our progressive three-stage optimization paradigm establishes a new state-of-the-art. Notably, our visual-level feature integration proves significantly superior to conventional textual prompt injection techniques. We believe SportsGrounder provides a robust and scalable new paradigm for object-centric multimodal understanding in complex, real-world environments.




\begin{acks}
This work was supported by the Zhejiang Provincial Natural Science Foundation of China (No. LZ24F030005), Fundamental Research Funds for the Central Universities (No. 226-2025-00167), National Natural Science Foundation of China (No. 62576308), and Research Fund for International Scientists of National Natural Science Foundation of China (72350710798).
\end{acks}


\bibliographystyle{ACM-Reference-Format}
\balance
\bibliography{sample-base}

\appendix


\end{document}